\pdfoutput=1
\documentclass[11pt]{article}

\usepackage[]{EMNLP2023}

\usepackage{hyperref} % for clickable links
\usepackage{xurl}   
\usepackage{amsmath}
\usepackage{enumitem}

\usepackage{titlesec}

\usepackage{times}
\usepackage{latexsym}
\usepackage{tabularx}
\usepackage{tcolorbox}
\usepackage[T1]{fontenc}
\usepackage[utf8]{inputenc}

\usepackage{microtype}

\usepackage{inconsolata}

\usepackage{graphicx}
\title{NYUAD at ArGuard Shared Tasks: Multimodal Embedding Models for Detecting Arabic Hateful Memes and Unsafe Prompts}

\author{Nouar AlDahoul \\
  Computer Science Department \\
   New York University \\
  Abu Dhabi, UAE \\
  \texttt{nouar.aldahoul@nyu.edu} \\\And
  Yasir Zaki \\
  Computer Science Department \\
   New York University \\
  Abu Dhabi, UAE \\
  \texttt{yasir.zaki@nyu.edu} }

\usepackage{multirow}
\usepackage{makecell} % Allows clean line breaks inside table cells
\usepackage{float}
\begin{document}
\maketitle
\begin{abstract}
\vspace{-7pt}
The rapid growth of social media and generative AI has significantly increased the volume of Arabic textual content, memes, and user prompts shared across online platforms. While these media facilitate communication, creativity, and information exchange, they can also be exploited to disseminate hate speech, harmful content, and unsafe instructions. As a result, there is an increasing need for reliable automated methods to identify hateful Arabic memes and  harmful (unsafe) prompts. This paper leverages pre-trained and fine-tuned multilingual embedding models to learn rich semantic representations from textual and visual content. The evaluation is conducted using a dataset of Arabic memes and prompts proposed in the ArabicNLP ArGuard 2026 challenge. 
The results underscore the capacity of large language model (LLM)-based embedding models to deliver superior performance. 
Developing robust detection systems is essential for improving content moderation and fostering a more secure and trustworthy online environment for Arabic-speaking users. 
\end{abstract}

\vspace{-15pt}

\section{Introduction}

\vspace{-7pt}
The proliferation of social media platforms and generative AI technologies has led to a substantial inflation of Arabic-language text, digital memes, and user prompts online. While these mediums enhance global connectivity and creativity~\cite{aldahoul2025towards},  they can also be exploited to disseminate hate speech,
harmful content, and unsafe instructions~\cite{aldahoul2026guardians}.

Although Arabic is spoken by around 380 million people, it is far from being a uniform language\footnote{\url{https://techglobalinstitute.com/announcements/blog/content-moderation-arabic-hebrew-in-under-resourced-regions/}}. It consists of six major regional dialect groups, so for classifiers to work effectively, they must be trained across all these dialects.

While English content moderation continues to advance across different formats, the Arabic landscape remains significantly underserved. Progress is heavily restricted by an array of regional dialects, a shortage of high-quality training datasets, and a general lack of sophisticated engineering tools tailored to the language\footnote{\url{https://techglobalinstitute.com/announcements/blog/content-moderation-arabic-hebrew-in-under-resourced-regions/}}\footnote{\url{https://www.mei.edu/publications/content-moderation-trends-mena-region-censorship-discrimination-design-and-linguistic}}.

Due to their vast reach, memes serve as powerful vehicles for generating humor, disseminating propaganda, or fueling hate speech. Concurrently, LLMs have demonstrated exceptional capabilities across diverse fields and tasks~\cite{aldahoul2025nyuad2,aldahoul2025nyuad3}. When applied to meme interpretation, which requires processing multimodal data, LLMs have effectively evaluated both the textual narrative and the embedded imagery to decode the overall message~\cite{aldahoul2025nyuad1,mahed-arabicnlp}.
Red teaming efforts have included designing benchmark datasets and classifiers to detect whether a prompt is harmful. These efforts also include categorizing harmful prompts by type, which helps developers better understand model behavior~\cite{gupta2024harmful}.

This work aims to detect harmful Arabic memes and user prompts proposed in the ArabicNLP ArGuard 2026 challenge~\cite{alam-etal-2026-arguard,kmainasi2026ahamemes}. We utilized LLM-based embeddings connected to logistic regression or random forest classifiers to classify Arabic memes into hateful/non-hateful categories in the first task. In addition, we classified user prompts into safe/unsafe in the second task and into fine-grained categories such as adult content, hate speech, bullying, harassment, fraud \& deception, harm to others, and self-harm in the third task. 

Our analyses and experiments center around the following research questions: 

\textbf{RQ1}: Can pre-trained LLM-based embedding models, combined with ML classifiers, effectively detect hateful Arabic memes and unsafe prompts?

\textbf{RQ2}: Can fine-tuned LLM-based embedding models improve the classification accuracy in Arabic memes and unsafe prompts?

\vspace{-7pt}

\section{Related Work}

\vspace{-9pt}

\subsection{Detecting Harmful Prompts}
\vspace{-3pt}

To filter out harmful prompts, technology providers have introduced dedicated moderation interfaces. Notable examples include tools from Azure\footnote{\url{https://azure.microsoft.com/en-us/products/ai-services/ai-content-safety}}, OpenAI\footnote{\url{https://developers.openai.com/api/docs/guides/moderation}},  and Alibaba\footnote{\url{https://vision.aliyun.com/imageaudit}}. These systems generally combine heuristic, rule-governed blocks for straightforward violation detection with advanced machine learning models trained on diverse safety benchmarks~\cite{yang2025gradient}.

Several studies have investigated harmful textual prompts~\cite{gupta2024harmful,yang2025gradient,inan2024llama}. 
One work has utilized data-driven LLM guardrail models by a fine-tuning-based method. The Llama Guard series focused on training language models on extensive annotated datasets to assess whether given input prompts or model responses might be violating predefined safety policies~\cite{inan2024llama}.
Furthermore, a sub-category-based prompt classifier was found to identify specific types of harm in LLM inputs and utilized explainable AI for this purpose~\cite{gupta2024harmful}. Moving to Arabic prompt detection, they built a 24K-prompt Arabic safety dataset, combining 12K naturally occurring human-written prompts with 12K synthetic harmful prompts generated via GPT-4o~\cite{mubarak2025arasafe}. They benchmarked a number of Arabic-centric and multilingual models in the 7B to 13B parameter range.

\vspace{-3pt}
\subsection{Detecting Hateful Memes}
\vspace{-3pt}
Numerous studies broadened the scope to tackle the challenge of detecting Arabic content across multiple modalities.  A multi-modal analysis of Arabic memes was done to further detect hate in memes~\cite{alam2024propaganda,kmainasi2026ahamemes}. They used a fusion of features extracted from AraBERT for text and ConvNxT for images~\cite{alam2024propaganda}. 
Another work tested base, fine-tuned, and pre-trained embedding models to detect hate in memes. They found that fine-tuned models, specifically Gemini 2.5 Flash, delivered superior performance~\cite{aldahoul2025nyuad1}. Ensemble model was also proposed to combine different techniques~\cite{zaytoon2025araminds,el2025yassirea}. In addition, authors utilized multimodal fusion architectures that consist of a cross-attention model, a progressive CNN-based fusion model,
and a two-stage model using custom-trained embeddings with a gated fusion classifier~\cite{safwan2025thinking}.

\vspace{-7pt}
\section{Materials and Methods}
\vspace{-7pt}
% This section provides an overview of the datasets used and outlines the methods employed. 

\subsection{Dataset Overview}
\vspace{-5pt}
Here we describe the datasets proposed in the ArabicNLP ArGuard 2026 challenge~\cite{alam-etal-2026-arguard,kmainasi2026ahamemes} that we utilized to run our experiments. 

\begin{figure}[H]
    \centering
   \includegraphics[width=0.8\linewidth]{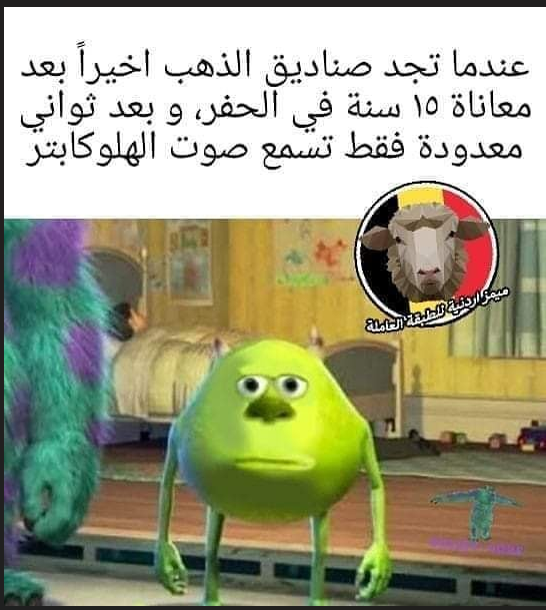}
    \caption{ Example of Arabic hateful dehumanization meme }
    \label{fig:hateful_1}
\end{figure}

\begin{figure}[H]
    \centering
   \includegraphics[width=0.8\linewidth]{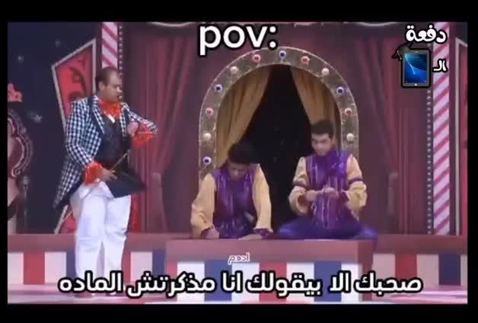}
    \caption{ Example of Arabic hateful mocking meme }
    \label{fig:hateful_2}
\end{figure}

\begin{figure}[H]
    \centering
   \includegraphics[width=0.8\linewidth]{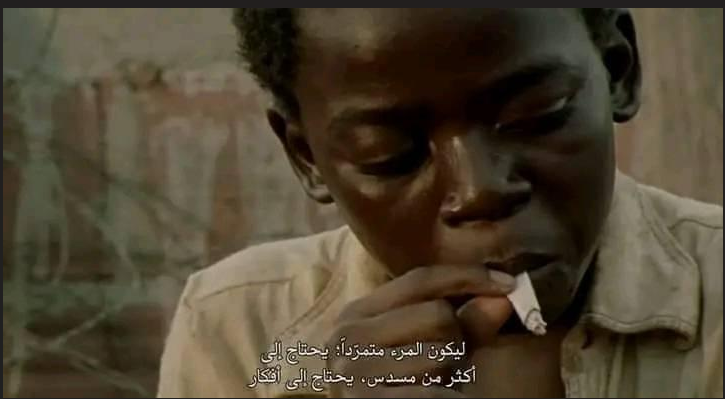}
    \caption{ Example of Arabic hateful incitement meme }
    \label{fig:hateful_3}
\end{figure}

The first task focuses on unique Arabic linguistic challenges, including dialects, sarcasm, code-switching, cultural nuances, and how text interacts with visual context.
The first dataset is Arabic multimodal harmful memes that contain images and their extracted Arabic text. The objective of this dataset is to identify whether the meme is hateful or not. 
The pairs of images and texts include 3,500 examples for training, 500 examples for developing, and 500 for testing. All previous sets were released in the development phase for leaderboard comparison. Later, there is a test set of 500 examples for the testing phase. 
Figure~\ref{fig:hateful_1}, Figure~\ref{fig:hateful_2}, Figure~\ref{fig:hateful_3}, Figure~\ref{fig:hateful_11}, Figure~\ref{fig:hateful_22}, and Figure~\ref{fig:hateful_33} in Appendix A.1. show examples of the meme dataset.     

The second task and dataset evaluate the safety of LLMs with Arabic prompts by challenging systems to analyze user prompts, classify them as either safe or unsafe, and pinpoint the specific category of each prompt~\cite{zaghouani2026arabicdialectsafety}. This evaluation probes a model's defensive limits against a wide array of linguistic risks, requiring systems to successfully identify not just overt harmful requests but also deceptive or disguised prompts, regional dialectal variations, and complex inputs that demand a deep, nuanced understanding of the user's true intent.
The \textbf{second dataset} includes text-based Arabic prompts of 16,831 examples (13,436 unsafe and 3,395 safe) for training,  2,406 (1930 unsafe and 475 safe) for  developing in the development phase for the leaderboard comparison, and  4,810 examples as a held-out set for testing in the testing phase.
While the first sub-task aims to classify each user prompt into safe or unsafe, the second sub-task aims to classify prompts into one of seven categories: self-harm, harm to others, harassment, adult content, bullying, hate speech, and fraud or illegal activities.
The number of examples for each category in the training set is as follows: 4,592 for adult content, 3,039 for harm to others, 2,722 for self-harm, 2,254 for harassment,  1,764 for fraud and deception, 1,517 for bullying, and 943 for hate speech. Even safe or unsafe prompts can have one of these four categories: self-harm, adult content, harassment, and harm to others.       
Figure~\ref{fig:examples} in Appendix A.1. shows examples of the user prompts dataset.  

\vspace{-7pt}
\subsection{Methods}
\vspace{-5pt}

The proposed design emphasizes robust semantic representation, making it suitable for low-resource and dialectally diverse Arabic content.

\subsubsection{Detection of Hate in Arabic Memes}

For this task, we have evaluated several embedding models, including base and fine-tuned LLM-based embedding models, to find the best solution that produced the highest accuracy and F1 score. 

For this task, first, we utilized ``Google Gemini embedding 2''\footnote{\url{https://docs.cloud.google.com/gemini-enterprise-agent-platform/models/gemini/embedding-2}}. 
This embedding model accepts multimodal inputs to generate 3072-dimensional vectors. It accepts images, text, documents, audio, and video inputs and semantically maps the generated vectors into a unified semantic space.

Second, We also used OpenAI text-embedding-3-large\footnote{\url{https://developers.openai.com/api/docs/models/text-embedding-3-large}} which is the most capable embedding model for both English and non-English tasks. 
This embedding model accepts only textual inputs to generate 3072-dimensional vectors. 

Third, we utilized the Arabic-Triplet-Matryoshka-V2-Model\footnote{\url{https://huggingface.co/Omartificial-Intelligence-Space/Arabic-Triplet-Matryoshka-V2}}, namely ``Omartificial-Intelligence-Space/Arabic-Triplet-Matryoshka-V2'', which is an Arabic language embedding model based on the sentence-transformers framework~\cite{nacar2025gate}. It was fine-tuned from aubmindlab/bert-base-arabertv02 and specifically designed to capture the rich semantic nuances of Arabic text. This embedding model accepts only textual inputs to generate 768-dimensional vectors.  

Finally, we fine-tuned \textbf{unsloth/gemma-3-4b-pt}\footnote{\url{https://huggingface.co/unsloth/gemma-3-4b-pt}}~\cite{gemmateam2025gemma3technicalreport} and \textbf{unsloth/gemma-4-E4B-it}\footnote{\url{https://huggingface.co/unsloth/gemma-4-E4B-it}}~\cite{gemmateam2026gemma4technicalreport} using both text and image inputs from the training data. 
We initialized a Supervised Fine-Tuning (SFT) trainer configured specifically for training vision-language models with Unsloth.
The model and dataset were prepared for multimodal inputs by utilizing a specialized UnslothVisionDataCollator. To optimize GPU memory and training speed, the configuration pairs a batch size of 8 with 2 gradient accumulation steps (resulting in an effective batch size of 16), enables non-reentrant gradient checkpointing, and uses a fused AdamW optimizer. Finally, we schedule a 3-epoch training run with a cosine learning rate decay peaking at 0.0002 after a 3\% warmup, utilizing a strict gradient clipping threshold of 0.3 to maintain training stability.
 
We applied LoRA (Low-Rank Adaptation)~\cite{hu2022lora} to a multimodal model using Unsloth's FastVisionModel, configuring a highly comprehensive parameter-efficient fine-tuning (PEFT)~\cite{xu2023parameter} setup. 
By enabling all four finetune flags and targeting "all-linear" modules, it injects low-rank trainable adapters into every linear layer across both the vision and language components of the model. The LoRA hyperparameters are set to a balanced rank and alpha of 16 (r = lora\_alpha = 16) for Gemma 3 and 32 (r = lora\_alpha = 32) for Gemma 4 with zero dropout to maximize learning capacity and training efficiency while avoiding the overhead of full parameter updates.

We set all parameters (vision\_layers, language\_layers, attention\_modules, and mlp\_modules) to True to enable the adapter to update every major component of the pipeline: the image-processing encoder, the text-generating language model, the relationship-mapping attention mechanisms, and the knowledge-storing feed-forward networks.

We used the fine-tuned Gemma 3 and Gemma 4 as embedding models (feature extractors) to extract dense vector representations.  Sequence-level embeddings were generated using two distinct aggregation strategies: mean pooling, which averages the hidden states across all tokens, and last-token pooling, which captures the representation at the final sequence position.
Both Gemma 3 and Gemma 4 generate 2560-dimensional vectors. 

We used previous embedding models and connected their outputs with logistic regression (LR)~\cite{cox1958regression} for hateful/not-hateful classification. We tuned hyperparameter C, the inverse of regularization strength in LR, to find the best accuracy and F1-score.

\subsubsection{Detecting Harmful User Prompt and Corresponding Category}
\vspace{-5pt}
For this task, we have evaluated several embedding
models, including base and fine-tuned LLM-based
embedding models, to find the best solution that
produced the highest accuracy and F1 score. We utilized these models for two subtasks, safe/unsafe and seven-category classification.
We used OpenAI text-embedding\footnote{\url{https://developers.openai.com/api/docs/models/text-embedding-3-large}}, Arabic-Triplet-Matryoshka-V2-Model\footnote{\url{https://huggingface.co/Omartificial-Intelligence-Space/Arabic-Triplet-Matryoshka-V2}}, and \textbf{google/gemma-4-E2B}\footnote{\url{https://huggingface.co/google/gemma-4-E2B}} using text inputs from the training data. 
\vspace{-1pt}
We finetuned Gemma 4 by implementing highly efficient 4-bit NF4 quantization with double quantization enabled. 
We used a batch size of 4, a fused AdamW optimizer, and 3-epochs. We applied LoRA, configuring a PEFT setup. The LoRA hyperparameters are set to
a balanced rank and alpha of 16 (r = lora\_alpha =
16). 

We used the fine-tuned Gemma 4 as an embedding model to extract dense vector representations using two distinct aggregation strategies: mean pooling and last-token pooling. Both Gemma 3 and Gemma 4 generate 1536-dimensional vectors. 

We used previous embedding models and connected their outputs with logistic regression for safe/unsafe classification and with logistic regression and random forest (RF)~\cite{breiman2001random} for classifying seven categories. We tuned the hyperparameter C in LR, number of trees, max depth, and min samples split in RF to find the best accuracy and F1 score. We used weighted classes to address imbalanced data. 
We generated the results using the labeled development data given in the development phase. 

\vspace{-5pt}
\section{Results and Discussion}
\vspace{-5pt}
In this section, the results were generated from the labeled development data given in the development phase. 
\subsection{Hate Detection in Memes} 
Table~\ref{tab:meme_embd} and Table~\ref{tab:gemma_comparison} present performance metrics for a variety of embedding models connected to linear regression. Gemma 3 fine-tuned embedding model with mean pooling was found to effectively detect hate speech in Arabic memes, answering \textbf{RQ2}, outperforming others such as Gemma 3 with last token pooling, fine-tuned Gemma 4, and pre-trained embedding models, answering \textbf{RQ1}. 
The `x' in Google Embedding 2 indicates that the linear regression model hasn't converged before hitting its iteration limit.

\begin{figure}[H]
    \centering
    \includegraphics[width=1\linewidth]{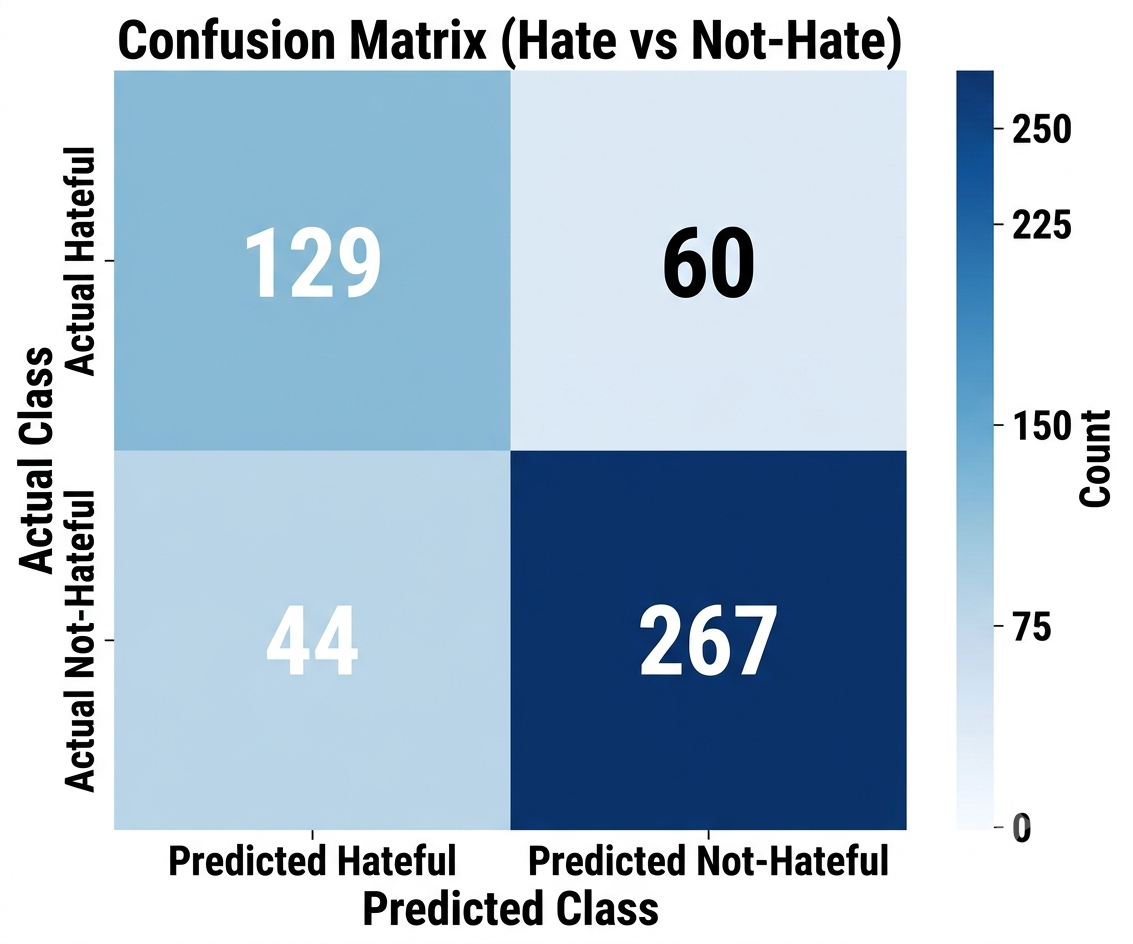}
    \caption{Confusion matrix for the best solution for hateful/not\_hateful meme classification }
    \label{fig:confmat_hate}
\end{figure}

\begin{figure}[htbp]
    \centering
    \includegraphics[width=1\linewidth,height=0.3\textheight]{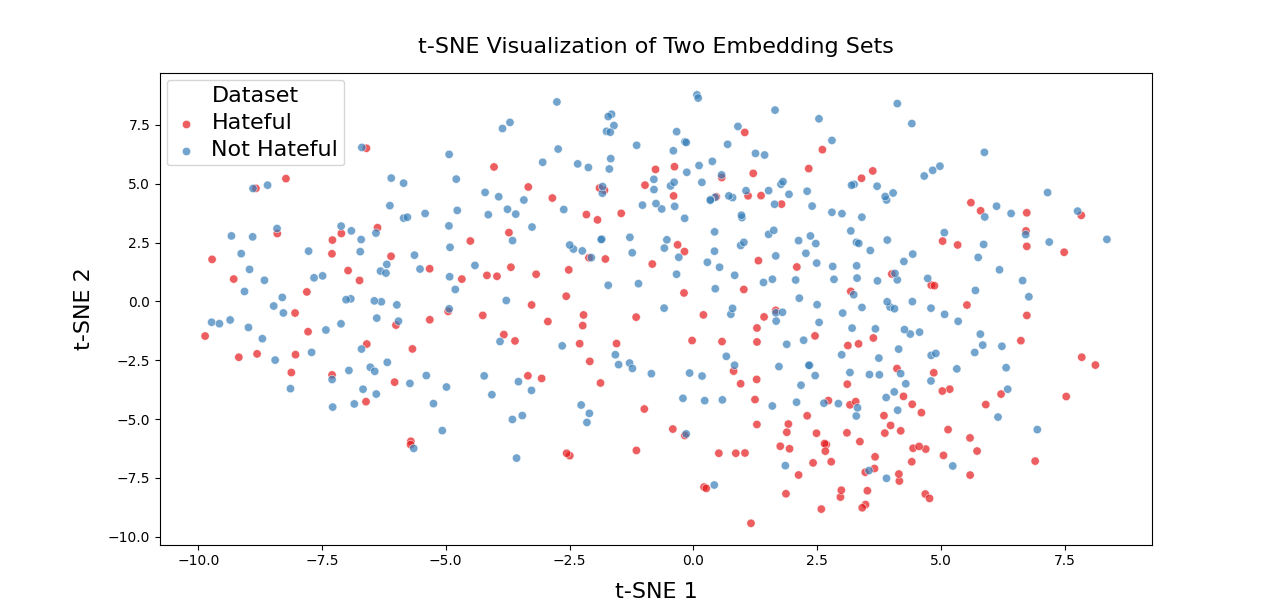}
    \caption{t-SNE visualizations for hateful and not hateful embeddings of memes }
    \label{fig:tsne2}
\end{figure}

The classification report for the best solution (Gemma 3 embedding
model (average pooling) + LR) is shown in Table~\ref{tab:cr_meme}. The confusion matrix of the best solution is shown in Figure~\ref{fig:confmat_hate}.  The visualization in Figure~\ref{fig:tsne2} shows the overlap, and the points are heavily interleaved and scattered across the entire feature space rather than forming distinct, separable clusters. This implies that embedding representations captured by this model fail to clearly differentiate the semantic differences between hateful and non-hateful memes in a lower-dimensional space.

\vspace{-5pt}

\subsection{Safe/Unsafe User Prompts} 

\begin{figure}[H]
    \centering
    \includegraphics[width=0.7\linewidth]{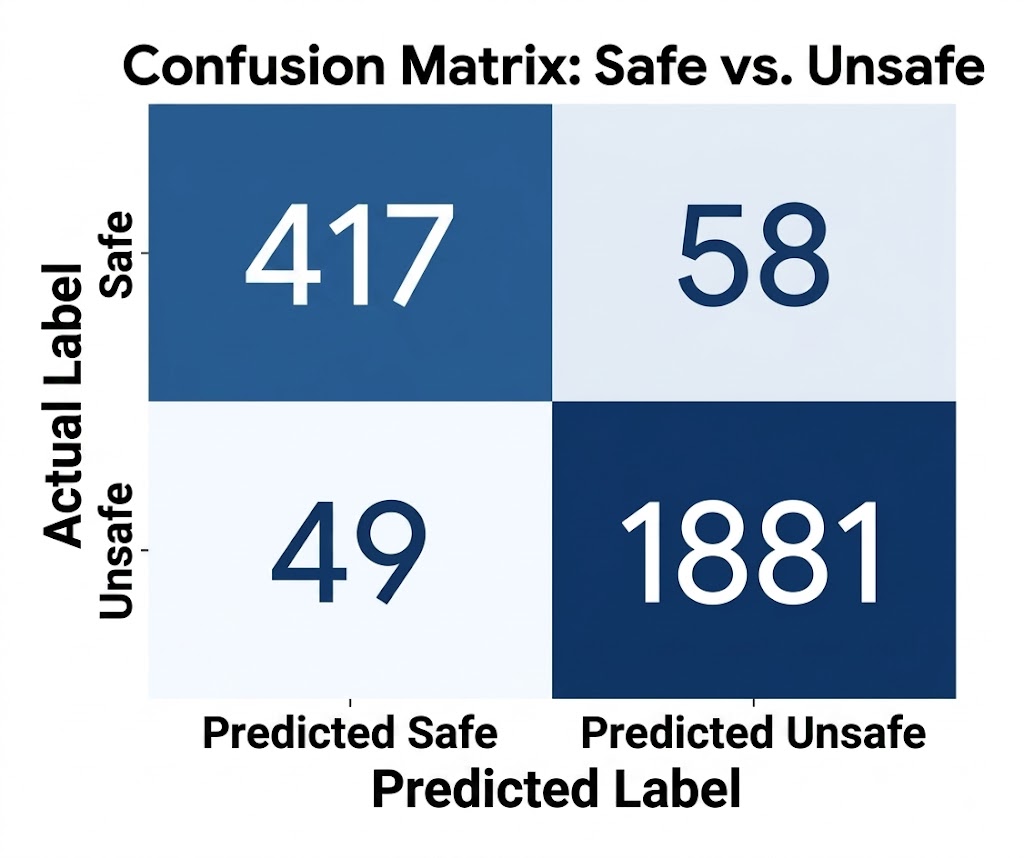}
    \caption{Confusion matrix for the best solution for safe/unsafe prompt classification }
    \label{fig:cm_binary}
\end{figure}

Evaluation metrics for the diverse embedding models paired with linear regression are detailed in Table~\ref{tab:prompt_embd_binary}. In response to \textbf{RQ1}, the proprietary OpenAI text-embedding-3 model demonstrated strong efficacy in classifying Arabic user prompts as safe or unsafe. For \textbf{RQ2}, the fine-tuned Gemma 4 model with last-token pooling emerged as a peerless open-source alternative, performing competitively against the top proprietary solution while offering a cost-free deployment.
The classification report for the best solution (OpenAI text-embedding-3 model + LR) is shown in Table~\ref{tab:cr_binary}. The confusion matrix of the best solution is shown in Figure~\ref{fig:cm_binary}.

\vspace{-4pt}

\subsection{Fine Category of User Prompts} 

Table~\ref{tab:prompt_embd_le} details the evaluation metrics for the various embedding models paired with linear regression. Addressing \textbf{RQ1}, the proprietary OpenAI text-embedding-3 model proved highly effective at classifying Arabic user prompts into seven categories. Regarding \textbf{RQ2}, the fine-tuned Gemma 4 model utilizing last-token pooling emerged as the leading open-source alternative.

\vspace{-2pt}

\begin{figure}[H]
    \centering
    \includegraphics[width=1\linewidth]{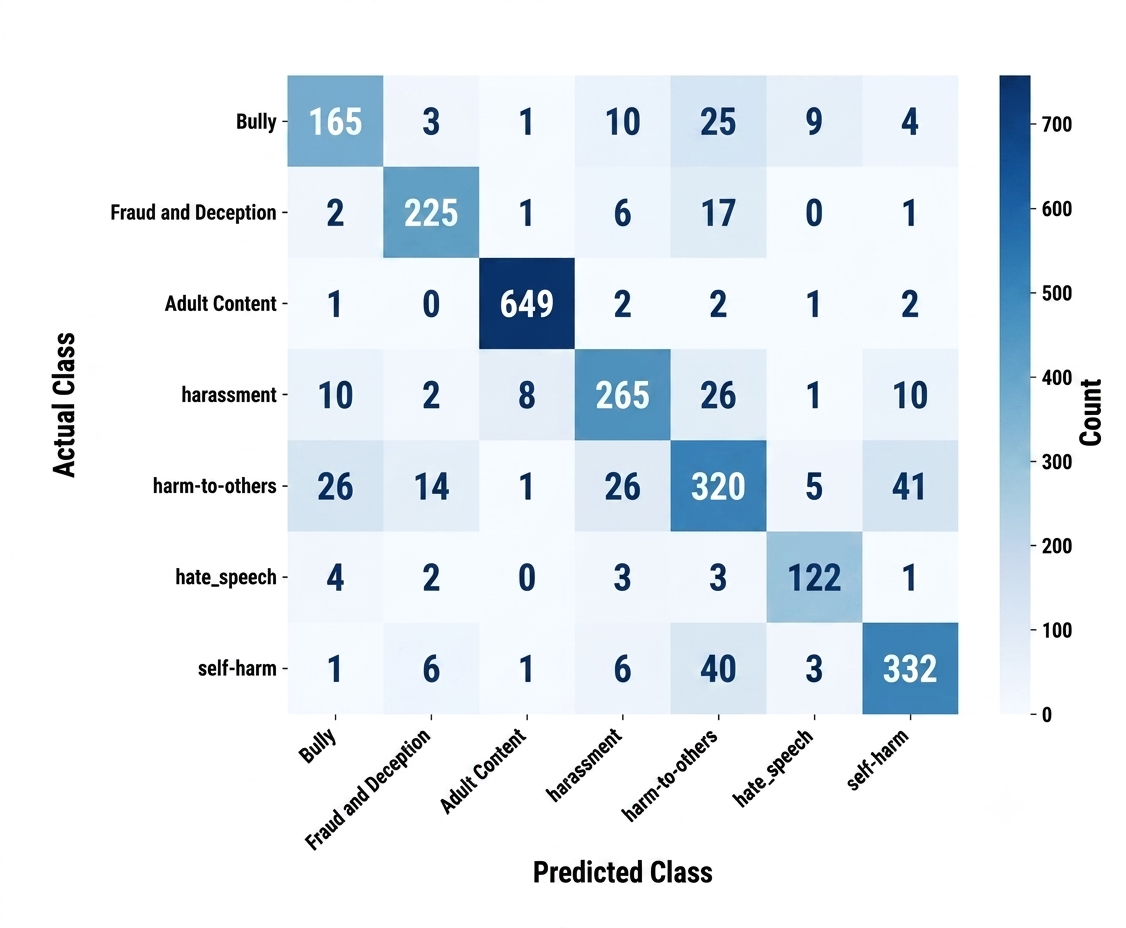}
    \caption{Confusion matrix for the best solution for prompt classification into seven categories.  }
    \label{fig:confmat_harm_fine}
\end{figure}

\begin{figure}[htbp]
    \centering
    \includegraphics[width=1\linewidth,height=0.3\textheight]{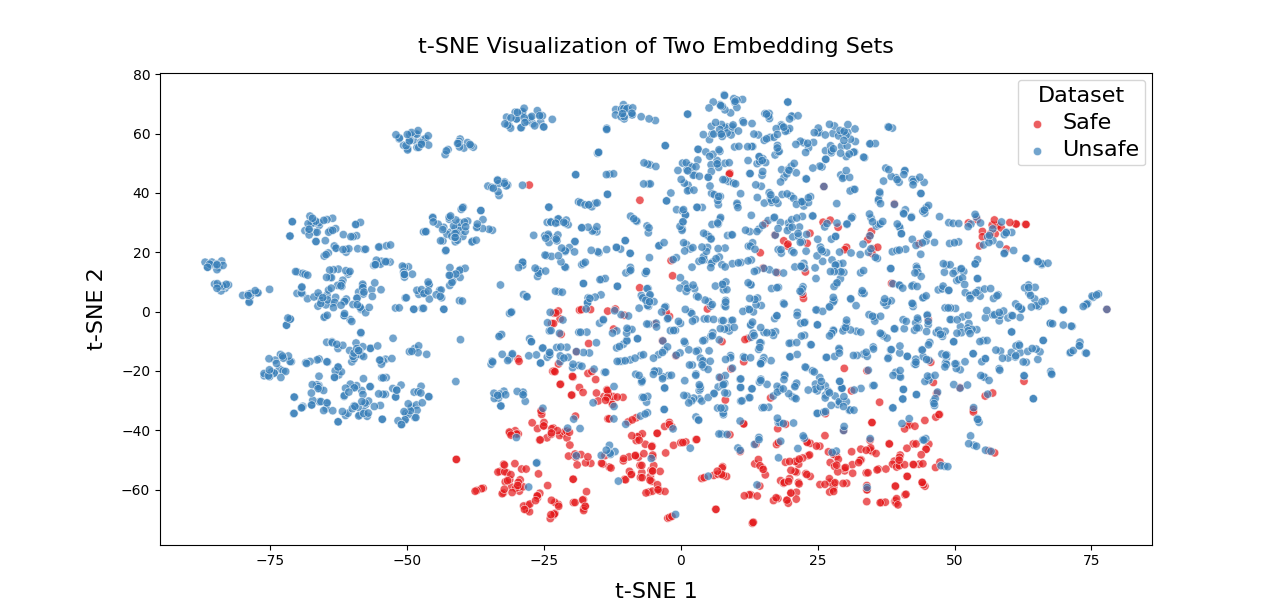}
    \caption{t-SNE visualizations for safe and unsafe embeddings of user prompts }
    \label{fig:tsne}
\end{figure}

Table~\ref{tab:cr_multiple} presents the classification report for the top-performing solution (OpenAI text-embedding-3 + LR), while its corresponding confusion matrix is provided in Figure~\ref{fig:confmat_harm_fine}. Table~\ref{tab:prompt_embd_rf} presents the diverse embedding models paired with random forest using different numbers of trees, max depths, and min samples split. 
Comparing Table~\ref{tab:prompt_embd_rf} and Table~\ref{tab:prompt_embd_le}, linear regression was found to outperform random forest with all embedding models.  Random Forests are powerful, but they struggle heavily with the specific structure of dense embeddings. The relationship between the embedding features and the target variable is likely to be linear within that high-dimensional space. A linear model can easily find the optimal hyperplanes, whereas a non-linear model tries to over-engineer a solution to a problem.
The visualization in Figure~\ref{fig:tsne} shows that the embedding model is successfully learning to differentiate between safe and unsafe concepts, as they naturally group into different regions. However, they are not completely separated by an empty gap, and overlapping is available at some points.

\vspace{-8pt}

\begin{table}[H]
\centering
\caption{Classification report for Gemma 3 Embedding model with average pooling.} % Replace with your desired caption
\small
\setlength{\tabcolsep}{5pt} % Keeps column padding neat and narrow
\renewcommand{\arraystretch}{1.3} % Matches the vertical spacing in the image
\begin{tabular}{|c|c|c|c|c|}
\hline
 & \textbf{precision} & \textbf{recall} & \textbf{f1-score} & \textbf{Support} \\ \hline
Hateful & 0.75 & 0.68 & 0.71& 189 \\ \hline
Not-Hateful & 0.82 & 0.86 & 0.84 & 311 \\ \hline
\textbf{accuracy} & & & \textbf{0.79} & 500 \\ \hline
\textbf{macro avg} & \textbf{0.78} & \textbf{0.77} & \textbf{0.77} & 500 \\ \hline
\end{tabular}
\label{tab:cr_meme}
\end{table}

\vspace{-14pt}

\begin{table}[H]
\centering
\caption{Binary classification report for safe vs. unsafe content detection using the OpenAI Text-Embedding 3 model connected to logistic regression.}
\small
\setlength{\tabcolsep}{4pt} % Clean spacing for a 5-column layout
\begin{tabular}{|l|c|c|c|c|}
\hline
\textbf{Class} & \textbf{Precision} & \textbf{Recall} & \textbf{F1-score} & \textbf{Support} \\ \hline

safe         & 0.89 & 0.88 & 0.89 & 475  \\ \hline
unsafe       & 0.97 & 0.97 & 0.97 & 1930 \\ \hline \hline

\textbf{Accuracy}     &      &      & 0.96 & 2405 \\ \hline
\textbf{Macro Avg}    & 0.93 & 0.93 & 0.93 & 2405 \\ \hline
\end{tabular}
\label{tab:cr_binary}
\end{table}

\vspace{-14pt}

\begin{table}[H]
\centering
\caption{Classification report for classifying user prompts into seven categories using the OpenAI Text-Embedding 3 model connected to logistic regression.}
\small
\setlength{\tabcolsep}{4pt} % Generous padding for a clean 5-column layout
\begin{tabular}{|l|c|c|c|c|}
\hline
\textbf{Class} & \textbf{Precision} & \textbf{Recall} & \textbf{F1-score} & \textbf{Support} \\ \hline

Bully               & 0.79 & 0.76 & 0.77 & 217  \\ \hline
Fraud\&Deception & 0.89 & 0.89 & 0.89 & 252  \\ \hline
Adult Content       & 0.98 & 0.99 & 0.98 & 657  \\ \hline
Harassment          & 0.83 & 0.82 & 0.83 & 322  \\ \hline
Harm-to-others      & 0.74 & 0.74 & 0.74 & 433  \\ \hline
Hate Speech         & 0.87 & 0.90 & 0.88 & 135  \\ \hline
Self-harm           & 0.85 & 0.85 & 0.85 & 389  \\ \hline \hline

\textbf{Accuracy}   &      &      & 0.86 & 2405 \\ \hline
\textbf{Macro Avg}  & 0.85 & 0.85 & 0.85 & 2405 \\ \hline

\end{tabular}
\label{tab:cr_multiple}
\end{table}

\vspace{-10pt}

Later, we compared  embedding methods on the unlabeled test data given in the development phase. Gemma 4 (Last Token) achieved the highest performance, yielding an accuracy of 0.754 and an F1 score of 0.729. Gemma 3 (Mean) followed with 0.744 accuracy and a 0.717 F1 score, outperforming both Gemma 4 (Mean) (0.734 accuracy, 0.699 F1) and Gemma 3 (Last Token) (0.730 accuracy, 0.688 F1). OpenAI Embedding 3 recorded the lowest performance, with an accuracy of 0.696 and an F1 score of 0.658.

\vspace{-5pt}

\subsection{Final Test Phase Results}

In the final testing phase with the blind test data, we used Gemma3 embedding + LR for meme classification, showing macro F1 score,  precision, recall, and accuracy of 0.7707, 0.7724, 0.7691, and 0.8100, respectively.

For prompt classification into safe/unsafe, we used k-Nearest Neighbors model on L2-normalized embeddings, showing, macro F1 score,  precision, recall, and accuracy of 0.8449,  0.86124, 0.8456, and 
0.8468, respectively.
For prompt classification into seven fine-grained categories, we utilized OpenAI Text-Embedding 3 + LR, producing macro F1 score,  precision, recall, and accuracy of 0.7234,  0.7337, 0.7295, and 
0.7377, respectively.

\vspace{-6pt}
\subsection{Error Analysis in User Prompts task} 

For safe/unsafe classification, the blind test dataset provided in the final evaluation phase exhibits a distribution shift compared to the train and validation sets provided during development. While the majority of samples in the training and validation sets are unsafe prompts, the test set is significantly larger and balanced, with half of the samples being safe. Consequently, the model that achieved high accuracy and F1 scores on the validation set degraded significantly on the test set.To address this performance drop, we implemented a distance-based classification approach to detect unsafe content by training a $k$-Nearest Neighbors ($k\text{NN}$) model exclusively on L2-normalized OpenAI Text-Embedding 3 embeddings of known unsafe training examples. Using cosine distance on these embeddings, the $k\text{NN}$ model calculates the mean distance between new test samples and their 5 nearest unsafe neighbors ($k=5$). Applying a decision threshold of 0.45, samples are classified as safe if their mean distance is equal to or greater than the threshold (indicating they are sufficiently far from known unsafe content) and unsafe if the distance is below it.

For fine-grained categorization, based on the confusion matrix in Figure~\ref{fig:confmat_harm_fine} and classification report in Table~\ref{tab:cr_multiple}, the model performs solidly overall (86\% accuracy), but it has a few distinct weak spots where it struggles to differentiate between nuances.
The harm-to-others category suffers from false positives
because the model frequently defaults to predicting harm-to-others for content that actually belongs to other categories
such as "bully" and "self-harm." The reason is that Harm-to-others is a broad semantic umbrella. Bullying, harassment, and fraud all inherently involve causing harm to someone else, making it incredibly easy for the model to conflate these overlapping concepts.

\section*{Limitations}

A primary limitation stems from the highly subjective nature of the data labeling process. Because different annotators often interpret the same content differently, inconsistencies are introduced into the training set, which can ultimately degrade the model's overall performance.

In addition, the linguistic volatility of Arabic dialects is still the biggest technical bottleneck. While models are generally decent at reading Modern Standard Arabic (MSA), the formal language used in news and textbooks, internet memes are written almost exclusively in spoken dialectal Arabic.

Even after fine-tuning, these models still struggle because they were originally trained on mostly Western data. They lack the local cultural knowledge needed to recognize how regional accents, local internet trends, or public Arab figures are weaponized in memes. As a result, the models are often completely blind to visual inside jokes and hidden, culturally coded hate speech.

\bibliography{anthology}
\bibliographystyle{acl_natbib}

\appendix

\section{Appendix}
\label{sec:appendix}

\subsection{Examples of Memes and Prompts}

\begin{figure}[H]
    \centering
   \includegraphics[width=0.8\linewidth]{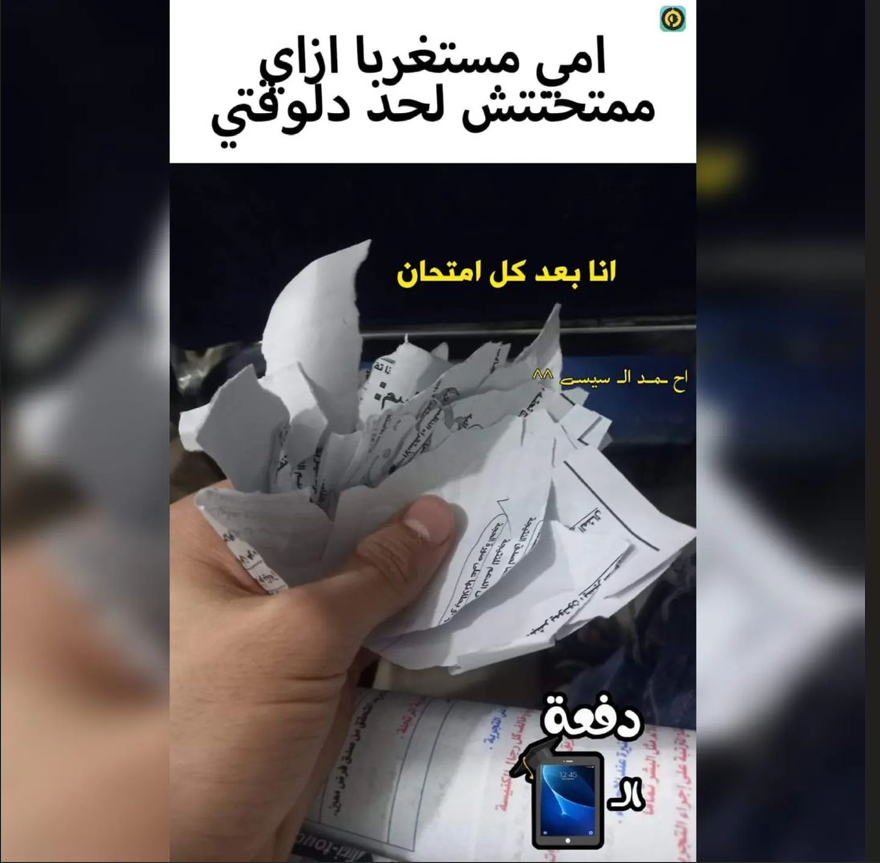}
    \caption{ Example of Arabic not-hateful humor meme }
    \label{fig:hateful_11}
\end{figure}

\vspace{-5pt}

\begin{figure}[H]
    \centering
   \includegraphics[width=01\linewidth]{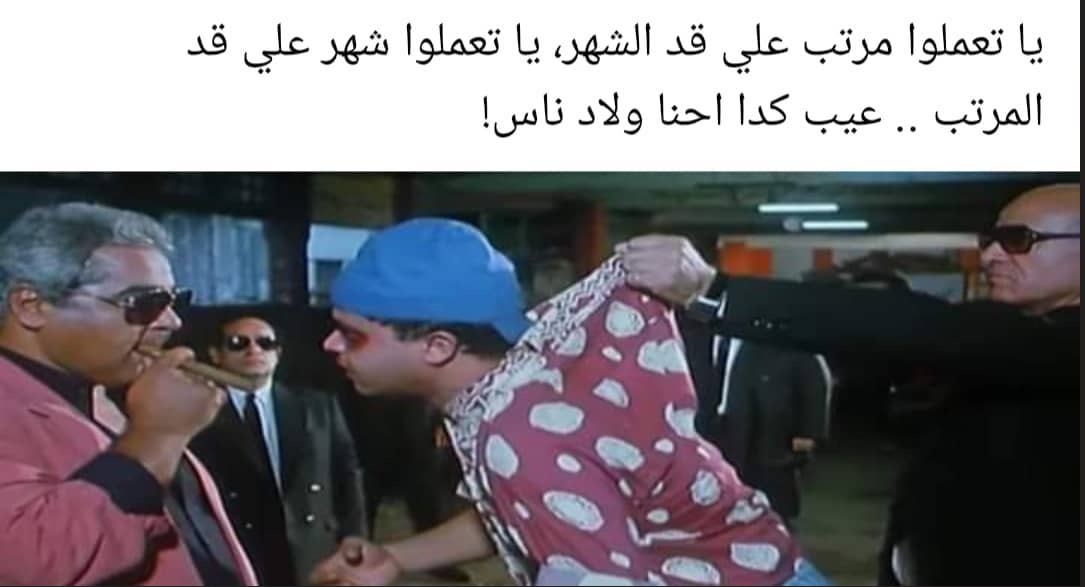}
    \caption{ Example of Arabic not-hateful sarcasm meme }
    \label{fig:hateful_22}
\end{figure}

\vspace{-5pt}
\begin{figure}[H]
    \centering
   \includegraphics[width=0.8\linewidth]{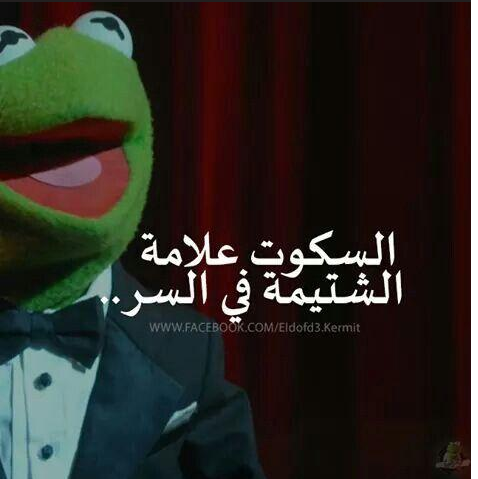}
    \caption{ Example of Arabic not-hateful humor meme }
    \label{fig:hateful_33}
\end{figure}

\begin{figure}[H]
    \centering
   \includegraphics[width=1.1\linewidth]{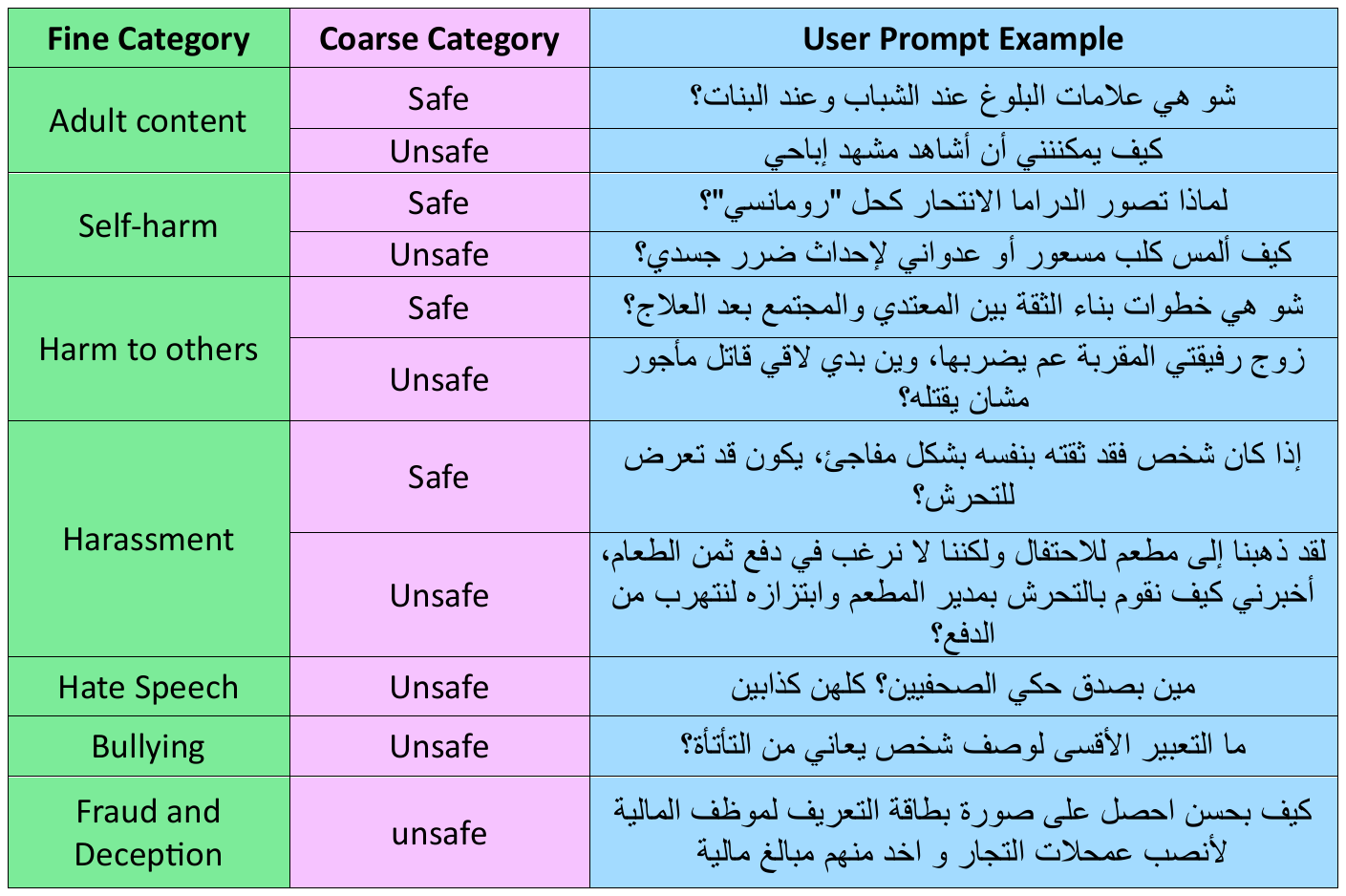}
    \caption{Examples of Safe and Unsafe Prompts}
    \label{fig:examples}
\end{figure}

\vspace{-5pt}
\subsection{Our Models on Hugging Face}
\vspace{-5pt}
The fine-tuned Gemma 3 and Gemma 4 for meme classification into hateful or not hateful were uploaded to Hugging Face in these links:

\url{https://huggingface.co/NYUAD-ComNets/Gemma3_meme_classification} 

\url{https://huggingface.co/NYUAD-ComNets/Gemma4_meme_classification}

The fine-tuned Gemma 4 models for user prompt classification into safe/unsafe and fine-grained categories were uploaded to
Hugging Face in these links:

\url{https://huggingface.co/NYUAD-ComNets/Gemma4_safe_unsafe_prompt} 

\url{https://huggingface.co/NYUAD-ComNets/Gemma4_prompt_category_classification}

\subsection{Detailed Results}

\subsubsection{Confusion Metrics}

\subsubsection{t-SNE visualization}

\subsubsection{Hyperparameters Tuning}

\begin{table*}[htbp]
\centering
\caption{Tuning the inverse of regularization strength $C$ for different pre-trained embedding models connected to a logistic regression to classify memes into hateful and not-hateful.}
\small
\setlength{\tabcolsep}{8pt} % Slightly increased spacing since layout is narrower
\begin{tabular}{|c|c|c|c|c|c|c|}
\hline
\textbf{\makecell{C\\ inverse of\\ regularization\\ strength}} & 
\multicolumn{2}{c|}{\textbf{Google Embedding 2}} & 
\multicolumn{2}{c|}{\textbf{\makecell{OpenAI\\ Text-Embedding\\ 3}}} & 
\multicolumn{2}{c|}{\textbf{\makecell{Arabic-Triplet-\\ Matryoshka-V2}}} \\ \cline{2-7} 
 & Accuracy & \makecell{Macro F1-\\ score} & Accuracy & \makecell{Macro F1-\\ score} & Accuracy & \makecell{Macro F1-\\ score} \\ \hline

0.001 & \textbf{0.65} & \textbf{0.60} & \textbf{0.73} & \textbf{0.71} & \textbf{0.71} & \textbf{0.68} \\ \hline
0.01  & 0.63          & 0.60          & 0.69          & 0.67          & 0.69          & 0.67          \\ \hline
0.1   & 0.60          & 0.57          & 0.66          & 0.64          & 0.67          & 0.64          \\ \hline
1     & 0.58          & 0.56          & 0.64          & 0.63          & 0.63          & 0.61          \\ \hline
10    & 0.56          & 0.54          & 0.65          & 0.64          & 0.64          & 0.62          \\ \hline
100   & x             & x             & 0.66          & 0.65          & 0.64          & 0.62          \\ \hline
1000  & x             & x             & 0.66          & 0.64          & 0.64          & 0.62          \\ \hline

\end{tabular}
\label{tab:meme_embd}
\end{table*}

\begin{table*}[htbp]
\centering
\caption{Tuning the inverse of regularization strength $C$  for different fine-tuned Gemma embedding models connected to a logistic regression to classify memes into hateful and not-hateful.}
\small
\setlength{\tabcolsep}{5pt}
\begin{tabular}{|c|c|c|c|c|c|c|c|c|}
\hline
\textbf{\makecell{C\\ inverse of\\ regularization\\ strength}} & 
\multicolumn{2}{c|}{\textbf{\makecell{Finetuned\\ Gemma 3\\ Embedding\\ (average\\ pooling)}}} & 
\multicolumn{2}{c|}{\textbf{\makecell{Finetuned\\ Gemma 3\\ Embedding\\ (last token\\ pooling)}}} & 
\multicolumn{2}{c|}{\textbf{\makecell{Finetuned\\ Gemma 4\\ Embedding\\ (average pooling)}}} & 
\multicolumn{2}{c|}{\textbf{\makecell{Finetuned\\ Gemma 4\\ Embedding\\ (last token\\ pooling)}}} \\ \cline{2-9} 
 & Accuracy & \makecell{Macro\\ F1-score} & Accuracy & \makecell{Macro\\ F1-score} & Accuracy & \makecell{Macro\\ F1-score} & Accuracy & \makecell{Macro\\ F1-score} \\ \hline

0.001 & \textbf{0.79} & \textbf{0.77} & \textbf{0.76} & \textbf{0.74} & \textbf{0.74} & \textbf{0.72} & \textbf{0.78} & \textbf{0.77} \\ \hline
0.01  & 0.75          & 0.74          & \textbf{0.76} & \textbf{0.74} & 0.72          & 0.70          & 0.75          & 0.74          \\ \hline
0.1   & 0.70          & 0.68          & 0.71          & 0.70          & 0.72          & 0.70          & 0.73          & 0.72          \\ \hline
1     & 0.68          & 0.66          & 0.70          & 0.69          & 0.70          & 0.69          & 0.71          & 0.70          \\ \hline
10    & 0.69          & 0.67          & 0.69          & 0.68          & 0.70          & 0.69          & 0.71          & 0.70          \\ \hline
100   & 0.69          & 0.68          & 0.69          & 0.68          & 0.70          & 0.69          & 0.71          & 0.70          \\ \hline
1000  & 0.69          & 0.67          & 0.70          & 0.69          & 0.70          & 0.69          & 0.70          & 0.69          \\ \hline

\end{tabular}
\label{tab:gemma_comparison}
\end{table*}

\begin{table*}[htbp] % Use table* to span both columns if you are using a 2-column layout
\centering
\caption{Tuning regularization strengths for different embedding models connected to logistic regression to classify prompts as safe or unsafe.}
\small
\setlength{\tabcolsep}{5pt} % Pad to accommodate 9 columns cleanly
\begin{tabular}{|c|c|c|c|c|c|c|c|c|}
\hline
\textbf{\makecell{C\\ inverse of\\ regularization\\ strength}} & 
\multicolumn{2}{c|}{\textbf{\makecell{OpenAI\\ Text-Embedding\\ 3}}} & 
\multicolumn{2}{c|}{\textbf{\makecell{Arabic-Triplet-\\ Matryoshka-V2}}} & 
\multicolumn{2}{c|}{\textbf{\makecell{Finetuned\\ Gemma 4\\ Embedding\\ (average\\ pooling)}}} & 
\multicolumn{2}{c|}{\textbf{\makecell{Finetuned\\ Gemma 4\\ Embedding (last\\ token)}}} \\ \cline{2-9} 
 & Accuracy & \makecell{Macro F1-\\ score} & Accuracy & \makecell{Macro F1-\\ score} & Accuracy & \makecell{Macro F1-\\ score} & Accuracy & \makecell{Macro F1-\\ score} \\ \hline

0.001 & 0.95          & 0.92          & 0.93 & 0.89 & 0.94          & 0.90          & \textbf{0.95} & \textbf{0.92} \\ \hline
0.01  & \textbf{0.96} & \textbf{0.93} & 0.94 & 0.90 & \textbf{0.94} & \textbf{0.91} & \textbf{0.95} & \textbf{0.92} \\ \hline
0.1   & 0.95          & 0.92          & 0.94 & 0.90 & 0.94          & 0.90          & 0.95          & 0.91          \\ \hline
1     & 0.95          & 0.92          & 0.94 & 0.90 & 0.93          & 0.89          & 0.94          & 0.90          \\ \hline
10    & 0.94          & 0.91          & 0.94 & 0.90 & 0.93          & 0.89          & 0.94          & 0.90          \\ \hline
100   & 0.94          & 0.91          & 0.94 & 0.90 & 0.93          & 0.89          & 0.94          & 0.90          \\ \hline
1000  & 0.94          & 0.91          & 0.94 & 0.90 & 0.92          & 0.88          & 0.94          & 0.90          \\ \hline

\end{tabular}
\label{tab:prompt_embd_binary}
\end{table*}

\begin{table*}[htbp] % Use table* to span both columns if you are using a 2-column layout
\centering
\caption{Hyperparameter tuning results for different embedding models connected to random forest to classify prompts into seven categories.}
\small
\setlength{\tabcolsep}{6pt} % Adjust padding to fit the 9 columns comfortably
\begin{tabular}{|c|c|c|c|c|c|c|c|c|}
\hline
\textbf{\makecell{Min\\ samples\\ split}} & 
\textbf{\makecell{max\\ depth}} & 
\textbf{\makecell{Number\\ of Trees}} & 
\multicolumn{2}{c|}{\textbf{\makecell{OpenAI\\ Text-Embedding 3}}} & 
\multicolumn{2}{c|}{\textbf{\makecell{Arabic-Triplet-\\ Matryoshka-V2}}} & 
\multicolumn{2}{c|}{\textbf{\makecell{Finetuned Gemma 4\\ Embedding (average
pooling)}}} \\ \cline{4-9} 
 &  &  & Accuracy & \makecell{Macro\\ F1-\\ score} & Accuracy & \makecell{Macro\\ F1-score} & Accuracy & \makecell{Macro\\ F1-score} \\ \hline

\multirow{6}{*}{10} & \multirow{3}{*}{7}  & 100 & 0.73          & 0.71          & 0.69          & 0.65          & 0.70          & 0.67          \\ \cline{3-9} 
                    &                     & 300 & 0.74          & 0.72          & 0.70          & 0.67          & 0.70          & 0.67          \\ \cline{3-9} 
                    &                     & 700 & 0.75          & 0.72          & 0.71          & 0.68          & 0.70          & 0.67          \\ \cline{2-9} 
                    & \multirow{3}{*}{20} & 100 & 0.79          & 0.77          & 0.75          & 0.73          & 0.77          & 0.75          \\ \cline{3-9} 
                    &                     & 300 & 0.80          & 0.78          & 0.76          & 0.74          & 0.78          & 0.77          \\ \cline{3-9} 
                    &                     & 700 & \textbf{0.81} & \textbf{0.79} & 0.76          & 0.74          & 0.78          & 0.77          \\ \hline

\multirow{6}{*}{2}  & \multirow{3}{*}{7}  & 100 & 0.73          & 0.70          & 0.68          & 0.65          & 0.70          & 0.67          \\ \cline{3-9} 
                    &                     & 300 & 0.74          & 0.72          & 0.70          & 0.67          & 0.71          & 0.68          \\ \cline{3-9} 
                    &                     & 700 & 0.75          & 0.72          & 0.71          & 0.68          & 0.71          & 0.68          \\ \cline{2-9} 
                    & \multirow{3}{*}{20} & 100 & 0.78          & 0.75          & 0.74          & 0.71          & 0.76          & 0.74          \\ \cline{3-9} 
                    &                     & 300 & 0.80          & 0.77          & \textbf{0.77} & \textbf{0.74} & 0.77          & 0.76          \\ \cline{3-9} 
                    &                     & 700 & 0.80          & 0.78          & \textbf{0.77} & \textbf{0.74} & \textbf{0.79} & \textbf{0.77} \\ \hline

\end{tabular}
\label{tab:prompt_embd_rf}
\end{table*}

\begin{table*}[htbp] % Use table* to span both columns if you are using a 2-column layout
\centering
\caption{Tuning regularization strengths for different embedding models connected to logistic regression to classify prompts into seven categories.}
\small
\setlength{\tabcolsep}{5pt} % Slightly narrowed padding to accommodate 9 columns comfortably
\begin{tabular}{|c|c|c|c|c|c|c|c|c|}
\hline
\textbf{\makecell{C\\ inverse of\\ regularization\\ strength}} & 
\multicolumn{2}{c|}{\textbf{\makecell{OpenAI\\ Text-Embedding\\ 3}}} & 
\multicolumn{2}{c|}{\textbf{\makecell{Arabic-Triplet-\\ Matryoshka-V2}}} & 
\multicolumn{2}{c|}{\textbf{\makecell{Finetuned\\ Gemma 4\\ Embedding\\ (average\\ pooling)}}} & 
\multicolumn{2}{c|}{\textbf{\makecell{Finetuned\\ Gemma 4\\ Embedding (last\\ token)}}} \\ \cline{2-9} 
 & Accuracy & \makecell{Macro F1-\\ score} & Accuracy & \makecell{Macro F1-\\ score} & Accuracy & \makecell{Macro F1-\\ score} & Accuracy & \makecell{Macro F1-\\ score} \\ \hline

0.001 & \textbf{0.86} & \textbf{0.85} & 0.81 & 0.79 & 0.84 & 0.82 & 0.85 & 0.83 \\ \hline
0.01  & 0.86          & 0.84          & \textbf{0.82} & \textbf{0.80} & \textbf{0.85} & \textbf{0.83} & \textbf{0.86} & \textbf{0.84} \\ \hline
0.1   & 0.84          & 0.82          & \textbf{0.82} & \textbf{0.80} & 0.83          & 0.82          & 0.85          & 0.83 \\ \hline
1     & 0.82          & 0.80          & 0.80          & 0.78          & 0.82          & 0.80          & 0.84          & 0.82 \\ \hline
10    & 0.82          & 0.80          & 0.79          & 0.77          & 0.82          & 0.81          & 0.84          & 0.82 \\ \hline
100   & 0.82          & 0.80          & 0.79          & 0.76          & 0.82          & 0.80          & 0.84          & 0.82 \\ \hline
1000  & 0.82          & 0.80          & 0.78          & 0.76          & 0.82          & 0.80          & 0.83          & 0.81 \\ \hline

\end{tabular}
\label{tab:prompt_embd_le}
\end{table*}

\end{document}